# AUTOMATIC CLASSIFICATION OF BENGALI SENTENCES BASED ON SENSE DEFINITIONS PRESENT IN BENGALI WORDNET

Alok Ranjan Pal[1], Diganta Saha[2] and Niladri Sekhar Dash[3]

[1]Dept. of Computer Science and Eng., College of Engineering and Management, Kolaghat
[2]Dept. of Computer Science and Eng., Jadavpur University, Kolkata
[3]Linguistic Research Unit, Indian Statistical Institute, Kolkata

## ABSTRACT

*Based on the sense definition of words available in the Bengali WordNet, an attempt is made to classify the Bengali sentences automatically into different groups in accordance with their underlying senses. The input sentences are collected from 50 different categories of the Bengali text corpus developed in the TDIL project of the Govt. of India, while information about the different senses of particular ambiguous lexical item is collected from Bengali WordNet. In an experimental basis we have used Naive Bayes probabilistic model as a useful classifier of sentences. We have applied the algorithm over 1747 sentences that contain a particular Bengali lexical item which, because of its ambiguous nature, is able to trigger different senses that render sentences in different meanings. In our experiment we have achieved around 84% accurate result on the sense classification over the total input sentences. We have analyzed those residual sentences that did not comply with our experiment and did affect the results to note that in many cases, wrong syntactic structures and less semantic information are the main hurdles in semantic classification of sentences. The applicational relevance of this study is attested in automatic text classification, machine learning, information extraction, and word sense disambiguation.*

## KEYWORDS

## 1. INTRODUCTION

In all natural languages, there are a lot of words that denote different meanings based on the contexts of their use within texts. Since it is not easy to capture the actual intended meaning of a word in a piece of text, we need to apply Word Sense Disambiguation (WSD) [1-6] technique for identification of actual meaning of a word based on its distinct contextual environments. For example in English, the word 'goal' may denote several senses based on its use in different types of construction, such as *He scored a goal, It was his goal in life,* etc. Such words with multiple meanings are ambiguous in nature and they posit serious challenges in understanding a natural language text both by man and machine.

The act of identifying the most appropriate sense of an ambiguous word in a particular syntactic context is known as WSD. A normal human being, due to her innate linguistic competence, is able to capture the actual contextual sense of an ambiguous word within a specific syntactic frame with the knowledgebase triggered from various intra- and extra-linguistic environments. Since a machine does not possess such capacities and competence, it requires some predefined rules or statistical methods to do this job successfully.

Normally, two types of learning procedure are used for WSD. The first one is Supervised Learning, where a learning set is considered for the system to predict the actual meaning of an ambiguous word within a syntactic frame in which the specific meaning for that particular word is embedded. The system tries to capture contextual meaning of the ambiguous word based on that defined learning set. The other one is Unsupervised Learning where dictionary information (i.e., glosses) of the ambiguous word is used to do the same task. In most cases, since digital dictionaries with information of possible sense range of words are not available, the system depends on on-line dictionaries like WordNet [7-13] or SenseNet.

Adopting the technique used in Unsupervised Learning, we have used the Naive Bayes [14] probabilistic measure to mark the sentence structure. Besides, we have a Bengali WordNet and a standard Bengali dictionary to capture the actual sense a word generates in a normal Bengali sentence.

The organization of the paper is as follows: in Section 2, we present a short review of some earlier works; in Section 3, we refer to the key features of Bengali morphology with reference to English; in Section 4, we present an overview of English and Bengali WordNet; in Section 5, we refer to the Bengali corpus we have used for our study; in Section 6, we explain the approach we have adopt for our work, in Section 7, we present the results and corresponding explanations; in Section 8, we present some close observations on our study, and in Section 9, we infer conclusion and redirect attention towards future direction of this research.

## 2. REVIEW OF EARLIER WORKS

WSD is perhaps one of the greatest open problems at lexical level of Natural Language Processing (Resnik and Yarowsky 1997). Several approaches have been established in different languages for assigning correct sense to an ambiguous word in a particular context (Gaizauskas 1997, Ide & Véronis 1998, Cucerzan, Schafer & Yarowsky 2002). Along with English, works have been done in many other languages like Dutch, Italian, Spanish, French, German, Japanese, Chinese, etc. (Xiaojie & Matsumoto 2003, Cañas, Valerio, Lalinde-Pulido, Carvalho & Arguedas 2003, Seo, Chung, Rim, Myaeng & Kim 2004, Liu, Scheuermann, Li & Zhu 2007, Kolte & Bhirud 2008, Navigli 2009, Nameh, Fakhrahmad & Jahromi (2011). And in most cases, they have achieved high level of accuracy in their works.

For Indian languages like Hindi, Bengali, Marathi, Tamil, Telugu, Malayalam, etc., effort for developing WSD system has not been much successful due to several reasons. One of the reasons is the morphological complexities of words of these languages. Words are morphologically so complex that there is no benchmark work in these languages (especially in Bengali). Keeping this reality in mind we have made an attempt to disambiguate word sense in Bengali. We believe this attempt will lead us to the destination through the tricky terrains of trial and error.

In essence, any WSD system typically involves two major tasks: (a) determining the different possible senses of an ambiguous word, and (b) assigning the word with its most appropriate sense in a particular context where it is used. The first task needs a Machine Readable Dictionary (MRD) to determine the different possible senses of an ambiguous word. At this moment, the most important sense repository used by the NLP community is the WordNet, which is being developed for all major languages of the world for language specific WSD task as well as for other linguistic works. The second task involves assigning each polysemic word with its appropriate sense in a particular context.

The WSD procedures so far used across languages may be classified into two broad types: (i) knowledge-based methods, and (ii) corpus-based methods. The knowledge-based methods obtain information from external knowledge sources, such as, Machine Readable Dictionaries (MRDs) and lexico-semantic ontologies. On the contrary, corpus-based methods gather information from the contexts of previously annotated instances (examples) of words. These methods extract knowledge from the examples applying some statistical or machine learning algorithms. When the examples are previously hand-tagged, the methods are called *supervised learning* and when the examples do not come with the sense labels they are called *unsupervised learning*.

## 2.1 Knowledge-based Methods

These methods do not depend on large amount of training materials as required in supervised methods. Knowledge-based methods can be classified further according to the type of resources they use: Machine-Readable Dictionaries (Lesk 1986); Thesauri (Yarowsky 1992); Computational Lexicon or Lexical Knowledgebase (Miller et al. 1990).

## 2.2 Corpus-based Methods

The corpus-based methods also resolute the sense through a classification model of example sentences. These methods involve two phases: *learning* and *classification*. The learning phase builds a sense classification model from the training examples and the classification phase applies this model to new instances (examples) for finding the sense.

## 2.3 Methods Based on Probabilistic Models

In recent times, we have come across cases where various statistics-based probabilistic models are being used to carry out the same task. The statistical methods evaluate a set of probabilistic parameters that express conditional probability of each lexical category given in a particular context. These parameters are then combined in order to assign the set of categories that maximizes its probability on new examples.

The Naive Bayes algorithm (Duda and Hart 1973) is the mostly used algorithm in this category, which uses the Bayes rule to find out the conditional probabilities of features in a given class. It has been used in many investigations of WSD task (Gale et al. 1992; Leacock et al. 1993; Pedersen and Bruce 1997; Escudero et al. 2000; Yuret 2004).

In addition to these, there are also some other methods that are used in different language for WSD task, such as, methods based on the similarity of examples (Schutze 1992), *k*-Nearest

Neighbour algorithm (Ng and Lee 1996), methods based on discursive properties (Gale et al. 1992; Yarowsky 1995), and methods based on discriminating rules (Rivest 1987), etc.

## 3. KEY FEATURES OF BENGALI MORPHOLOGY

In English, compared to Indic languages, most of the words have limited morphologically derived variants. Due to this factor it is comparatively easier to work on WSD in English as it does not pose serial problems to deal with varied forms. For instance, the verb *eat* in English has five conjugated (morphologically derived) forms only, namely, *eat, eats, ate, eaten,* and *eating*. On the other hand, most of the Indian languages (e.g., Hindi, Bengali, Odia, Konkani, Gujarati, Marathi, Punjabi, Tamil, Telugu, Kannada, Malayalam, etc.) are morphologically very rich, varied and productive. As a result of this, we can derive more than hundred conjugated verb forms from a single verb root. For instance, the Bengali verb খাওয়া (khāoyā) "to eat" has more than 150 conjugated forms including both *calit* (colloquial) and *sādhu* (chaste) forms, such as, খাই (khai), খাস (kkās), খাও (khāo), খায় (khāy), খান (khān), খাচ্ছি (khācchi), খাচ্ছিস (khācchis), খাচ্ছ (khāccha), খাচ্ছেন (khācchen), খাচ্ছে (khācche), খাইতেছি (khāitechi), খেয়েছি (kheyechi), খেয়েছ (kheyecha), খেয়েছিস (kheyechis), খেয়েছে (kheyeche), খেয়েছেন (kheyechen), খেলাম (khelam), খেলি (kheli), খেলে (khele), খেল (khela), খেলেন (khelen), খাব (khāba), খাবি (khābi), খাবে (khābe), খাবেন (khāben), খাচ্ছিলাম (khācchilām), খাচ্ছিলে (khācchile), খাচ্ছিল (khācchila), খাচ্ছিলেন (khācchilen), খাচ্ছিলি (khācchili), etc. (to mention a few).

While nominal and adjectival morphology in Bengali is light (in the sense that the number of derived forms from an adjective or a noun, is although quite large, in not up to the range of forms derived from a verb), the verbs are highly inflected. In general, nouns are inflected according to seven grammatical cases (nominative, accusative, instrumental, ablative, genitive, locative, and vocative), two numbers (singular and plural), a few determiners like, -টা (-ṭā), -টি (-ṭi), -খানা (-khānā), -খানি (-khāni), and a few emphatic markers, like -ই (-i) and -ও (-o), etc. The adjectives, on the other hand, are normally inflected with some primary and secondary adjectival suffixes denoting degree, quality, quantity, and similar other attributes. As a result, to build up a complete and robust system for WSD for all types of morphologically derived forms tagged with lexical information and semantic relations is a real challenge for a language like Bengali [15-25].

## 4. ENGLISH AND BENGALI WORDNET

The WordNet is a digital lexical resource, which organizes lexical information in terms of word meanings. It is a system for bringing together different lexical and semantic relations between words. In a language, a word may appear in more than one grammatical category and within that grammatical category it can have multiple senses. These categories and all senses are captured in the WordNet. WordNet supports the major grammatical categories, namely, Noun, Verb, Adjective, and Adverb. All words which express the same sense (same meaning) are grouped together to form a single entry in WordNet, called Synset (set of synonyms). Synsets are the basic building blocks of WordNet. It represents just one lexical concept for each entry. WordNet is developed to remove ambiguity in cases where a single word denotes more than one sense.

The English WordNet [26] available at present contains a large list of synsets in which there are 117097 unique nouns, 11,488 verbs, 22141 adjectives, and 4601 adverbs (Miller, Beckwith, Fellbaum, Gross, and Miller 1990, Miller 1993). The semantic relations for each grammatical category, as maintained in this WordNet, may be understood from the following diagrams (Fig. 1 and Fig. 2):

| Relation | Also called | Definition | Example |
|----------|-------------|------------|---------|
| Hypernym | Superordinate | From concepts to superordinates | $breakfast^1 \rightarrow meal^1$ |
| Hyponym | Subordinate | From concepts to subtypes | $meal^1 \rightarrow lunch^1$ |
| Member Meronym | Has-Member | From groups to their members | $faculty^2 \rightarrow professor^1$ |
| Has-Instance | | From concepts to instances of the concept | $composer^1 \rightarrow Bach^1$ |
| Instance | | From instances to their concepts | $Austen^1 \rightarrow author^1$ |
| Member Holonym | Member-Of | From members to their groups | $copilot^1 \rightarrow crew^1$ |
| Part Meronym | Has-Part | From wholes to parts | $table^2 \rightarrow leg^3$ |
| Part Holonym | Part-Of | From parts to wholes | $course^7 \rightarrow meal^1$ |
| Antonym | | Opposites | $leader^1 \rightarrow follower^1$ |

Figure 1. Noun Relations in English WordNet

| Relation | Definition | Example |
|----------|------------|---------|
| Hypernym | From events to superordinate events | $fly^9 \rightarrow travel^5$ |
| Troponym | From a verb (event) to a specific manner elaboration of that verb | $walk^1 \rightarrow stroll^1$ |
| Entails | From verbs (events) to the verbs (events) they entail | $snore^1 \rightarrow sleep^1$ |
| Antonym | Opposites | $increase^1 \Longleftrightarrow decrease^1$ |

Figure 2. Verb relations in English WordNet

The Bengali WordNet [27] is also a similar type of digital lexical resource, which aims at providing mostly semantic information for general conceptualization, machine learning and knowledge representation in Bengali (Dash 2012). It provides information about Bengali words from different angles and also gives the relationship(s) existing between words. The Bengali WordNet is being developed using expansion approach with the help of tools provided by Indian Institute of Technology (IIT) Bombay. In this WordNet, a user can search for a Bengali word and get its meaning. In addition, it gives the grammatical category namely, noun, verb, adjective or adverb of the word being searched. It is noted that a word may appear in more than one grammatical category and a particular grammatical category can have multiple senses. The WordNet also provides information for these categories and all senses for the word being searched.

Apart from the category for each sense, the following set of information for a Bengali word is presented in the WordNet:

(a)      Meaning of the word,
(b)      Example of use of the word,
(c)      Synonyms (words with similar meanings),
(d)      Part-of-speech,
(e)      Ontology (hierarchical semantic representation),

(f)      Semantic and lexical relations.

At present the Bengali WordNet contains 36534 words covering all major lexical categories, namely, noun, verb, adjective, and adverb.

# 5. THE BENGALI CORPUS

The Bengali corpus that is used in this work is developed under the TDIL (Technology Development for the Indian Languages) project, Govt. of India (Dash 2007). This corpus contains text samples from 85 text categories or subject domains like Physics, Chemistry, Mathematics, Agriculture, Botany, Child Literature, Mass Media, etc. (Table 1) covering 11,300 A4 pages, 271102 sentences and 3589220 non-tokenized words in their inflected and non-inflected forms. Among these total words there are 199245 tokens (i.e., distinct words) each of which appears in the corpus with different frequency of occurrence. For example while the word মাথা (māthā) "head" occurs 968 times, মাথায় (māthāy) "on head" occurs 729 times, মাথার (māthār) "of head" occurs 398 times followed by other inflected forms like মাথাতে (māthāte) "in head", মাথাটা (māthāṭā) "the head", মাথাটি (māthāṭi) "the head", মাথাগুলো (māthāgulo) "heads", মাথারা (māthārā) "heads", মাথাদের (māthāder) "to the heads", মাথারই (māthāri) "of head itself" with moderate frequency. This corpus is exhaustively used to extract sentences of a particular word required for our system as well as for validating the senses evoked by the word used in the sentences.

# 6. PROPOSED APPROACH

In the proposed approach, we have used Naive Bayes probabilistic model to classify the sentences based on some previously tagged learning sets. We have tested the efficiency of the algorithm over the Bengali corpus data stated above. In this approach we have used a sequence of steps (Fig. 3) to disambiguate the sense of māthā (head) – one of the most common ambiguous words in Bengali. The category-wise results are explained in results and evaluation section.

## 6.1 Text annotation

At first, all the sentences containing the word 'māthā' are extracted from the Bengali text corpus (Section 5). Total number of sentence counts: 1747. That means there are at least 1747 sentences in this particular corpus where the word মাথা (māthā) has been used in its non-inflected and non-compounded lemma form. However, since the sentences extracted from the corpus are not normalized adequately, these are passed through a series of manual normalization for (a) separation or detachment of punctuation marks like single quote, double quote, parenthesis, comma, etc. that are attached to words; (b) conversion of dissimilar fonts into similar ones; (c) removal of angular brackets, uneven spaces, broken lines, slashes, etc. from sentences; and (d) identification of sentence terminal markers (i.e., full stop, note of exclamation, and note of interrogation) that are used in written Bengali texts.

## 6.2 Stop word removal

The very next stage of our strategy was the removal of stop words. Based on traditional definition and argument, we have identified all **postpositions**, e.g., দিকে (dike) "towards", প্রতি (prati) "per", etc.; **conjunctions**, e.g., এবং (ebang) "and", কিন্তু (kintu) "but", etc.; **interjections**, e.g., বা! (bāh)

"well", আহা (āhā) "ah!", etc.; **pronouns**, e.g., আমি (āmi) "I", তুমি (tumi) "you", সে (se) "she" etc.; **some adjectives**, e.g., লাল (lāl) "red", ভাল (bhālo) "good", etc.; **some adverbs**, e.g., খুব (khub) "very", সত্যি (satyi) "really", etc.; **all articles**, e.g., একটি (ekṭi) "one", etc. and **proper nouns**, e.g., রাম (rām) "Ram", কলকাতা (kalkātā) "Calcutta", etc. as stop words in Bengali.

To identify stop words the first step was to measure frequency of use of individual words, which we assumed, would have helped us to identify stop words. However, since the term frequencies of stop words were either very high or very low, it was not possible to set a particular threshold value for filtering out stop words. So, in our case, stop words are manually tracked and tagged with the help of a standard Bengali dictionary.

## 6.3 Learning Procedure

As the proposed approach is based on supervised learning methodology, it is necessary to build up a strong learning set before testing a new data set. In our approach, we have therefore used three types of learning sets that are built up according to three different meanings of the ambiguous word মাথা (māthā) "head", which are collected from the Bengali WordNet. There are five types of dictionary definition (glosses) for the word with twenty five (25) varied senses in the WordNet, such as the followings:

i.        Category: Noun

Synonyms: মাথা, মস্তক, ললাট;

Concept: মাথার উপরের এবং সামনের অংশ

Example: রামের মাথায় আভা বিচ্ছুরিত হচ্ছে

ii.       Category: Noun

Synonyms: মাথা, মুণ্ড;

Concept: শরীরে গলার সামনের বা উপরের সেই গোলাকার অংশ যেখানে চোখ, কান, নাক, মুখ ইত্যাদি অঙ্গ থাকে এবং যার মধ্যে মস্তিষ্ক থাকে

Example: মাথায় আঘাত লাগার ফলে মানুষের প্রাণও যেতে পারে

iii.      Category: Noun

Synonyms: মাথা

Concept: শরীরের সেই অংশ যার মধ্যে মস্তিষ্ক থাকে

Example: মোহনের মাথায় চুল নেই

iv.      Category: Noun

Synonyms: মাথা

Concept: নৌকা বা জলযানের অগ্রভাগ

Example: তিনি বিশ্রাম নেওয়ার জন্য নৌকার মাথায় গিয়ে বসলেন

v.       Category: Noun

Synonyms: মাথা

Concept: কোনো উঁচু ভবন, মহল প্রভৃতির শিখর

Example: যে বাড়ির মাথায় চিল বানানো রয়েছে, আমি সেখানেই থাকি

It is observed from the categories that the first category represent first type of meaning (মাথা = মস্তক = ললাট), 2nd and 3rd category represent second type of meaning (মাথা = মুণ্ড), 4th and 5th category represent third type of meaning (অগ্রভাগ = শিখর = প্রান্তভাগ). Based on the information types we have built up three specific categories of senses of মাথা (māthā): (a) "মস্তক, ললাট", (b) "মাথা, মুণ্ড" and (c) "অগ্রভাগ, শিখর, প্রান্তভাগ". After this we have randomly chosen 55 sentences of each type of sense from the corpus to build the learning sets.

## 6.4 Modular representation of the proposed approach

In this proposed approach all the sentences containing মাথা (māthā) in the Bengali corpus are classified in three pre-defined categories using Naive Bayes (NB) classifier in the following manner (Fig. 3):

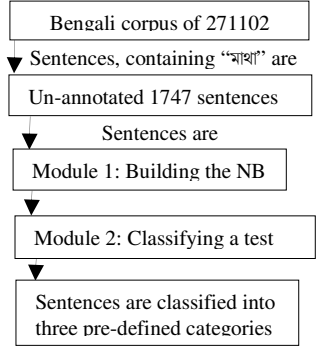

Figure 3. Overall procedure is represented graphically

### 6.4.1 Explanation of Module 1: Building NB model

In the NB model the following parameters are calculated based on the training documents:

- |V| = the number of vocabularies, means the total number of distinct words belong to all the training sentences.
- P($c_i$) = the priori probability of each class, means the number of sentences in a class / number of all the sentences.
- $n_i$ = the total number of word frequency of each class.
- P($w_i$ | $c_i$) = the conditional probability of keyword occurrence in a given class.

To avoid "zero frequency" problem, we have applied Laplace estimation by assuming a uniform distribution over all words, as-

P($w_i$ | $c_i$) = (Number of occurrences of each word in a given class + 1)/($n_i$ + |V|)

### 6.4.2 Explanation of Module 2: Classifying a test document

To classify a test document, the "posterior" probabilities, P($c_i$ | W) for each class is calculated, as-

$$P(c_i \mid W) = P(c_i) \text{ x } \sum_{j=1}^{|V|} P(w_j \mid c_i)$$

The highest value of probability categorizes the test document into the related classifier.

## 7. RESULTS AND CORRESPONDING EVALUATIONS

We performed testing operation over 271102 sentences of the Bengali corpus. As mentioned in section 6.1, this corpus consists of total 85 text categories of data sets like Agriculture, Botany, Child Literature, etc. in which there are 1747sentences containing the particular ambiguous word

মাথা (māthā). After annotation (Section 6.1), each individual sentence is passed through the Naive-Bayes model and the "posterior" probabilities, $P(c_i \mid W)$ for each sentence is evaluated. Greater probability represents the associated sense for that particular sentence.

The category-wise result is furnished in Table 1. The performance of our system is measured on Precision, Recall and F-Measure parameters, stated as-

Precision (P) = Number of instances, responded by the system/ total number of sentences present in the corpus containing মাথা (māthā).

Recall (R) = Number of instances matched with human decision/ total number of instances.

F-Measure (FM) = (2 * P * R) / (P + R).

Table 1. Performance analysis on the whole corpus.

| Category | Total no of sentence | মাথা | | ললাট/কপাল | | চূড়া/প্রান্ত | | Total right | Total wrong | P | R | FM |
|---|---|---|---|---|---|---|---|---|---|---|---|---|
| | | right | wrong | right | wrong | right | wrong | | | | | |
| Accountancy | 0 | 0 | 0 | 0 | 0 | 0 | 0 | 0 | 0 | 0 | NA | NA |
| Agriculture | 13 | 1 | 0 | 0 | 0 | 10 | 2 | 11 | 2 | 1 | 0.85 | 0.92 |
| Anthropology | 24 | 19 | 5 | 0 | 0 | 0 | 0 | 19 | 5 | 1 | 0.79 | 0.88 |
| Astrology | 6 | 4 | 0 | 2 | 0 | 0 | 0 | 6 | 0 | 1 | 1.00 | 1.00 |
| Astronomy | 3 | 1 | 1 | 1 | 0 | 0 | 0 | 2 | 1 | 1 | 0.67 | 0.80 |
| Banking | 1 | 0 | 0 | 0 | 0 | 0 | 1 | 0 | 1 | 1 | 0.00 | 0.00 |
| Biography | 50 | 24 | 3 | 17 | 0 | 6 | 0 | 47 | 3 | 1 | 0.94 | 0.97 |
| Botany | 3 | 2 | 0 | 0 | 0 | 1 | 0 | 3 | 0 | 1 | 1.00 | 1.00 |
| Business Math | 2 | 0 | 0 | 0 | 0 | 1 | 1 | 1 | 1 | 1 | 0.50 | 0.67 |
| Child Lit | 86 | 36 | 7 | 21 | 1 | 20 | 1 | 77 | 9 | 1 | 0.90 | 0.94 |
| Criticism | 21 | 5 | 0 | 7 | 2 | 5 | 2 | 17 | 4 | 1 | 0.81 | 0.89 |
| Dancing | 2 | 0 | 0 | 1 | 0 | 0 | 1 | 1 | 1 | 1 | 0.50 | 0.67 |
| Drawing | 3 | 1 | 1 | 1 | 0 | 0 | 0 | 2 | 1 | 1 | 0.67 | 0.80 |
| Economics | 16 | 0 | 0 | 3 | 0 | 12 | 1 | 15 | 1 | 1 | 0.94 | 0.97 |
| Education | 1 | 0 | 0 | 1 | 0 | 0 | 0 | 1 | 0 | 1 | 1.00 | 1.00 |
| Essay | 23 | 7 | 1 | 10 | 0 | 5 | 0 | 22 | 1 | 1 | 0.96 | 0.98 |
| Folk | 0 | 0 | 0 | 0 | 0 | 0 | 0 | 0 | 0 | 0 | NA | NA |
| GameSport | 60 | 34 | 4 | 13 | 0 | 9 | 0 | 56 | 4 | 1 | 0.93 | 0.97 |
| GenSc | 25 | 8 | 2 | 5 | 0 | 10 | 0 | 23 | 2 | 1 | 0.92 | 0.96 |
| Geology | 0 | 0 | 0 | 0 | 0 | 0 | 0 | 0 | 0 | 0 | NA | NA |
| HistoryWar | 2 | 1 | 0 | 1 | 0 | 0 | 0 | 2 | 0 | 1 | 1.00 | 1.00 |

| | | | | | | | | | | | |
|---|---|---|---|---|---|---|---|---|---|---|---|
| HomeSc | 38 | 12 | 1 | 3 | 0 | 21 | 1 | 36 | 2 | 1 | 0.95 | 0.97 |
| Humor | 33 | 6 | 3 | 16 | 0 | 8 | 0 | 30 | 3 | 1 | 0.91 | 0.95 |
| Journalism | 3 | 0 | 0 | 1 | 0 | 2 | 0 | 3 | 0 | 1 | 1.00 | 1.00 |
| Law&Order | 11 | 2 | 0 | 5 | 0 | 4 | 0 | 11 | 0 | 1 | 1.00 | 1.00 |
| Legislative | 6 | 3 | 3 | 0 | 0 | 0 | 0 | 3 | 3 | 1 | 0.50 | 0.67 |
| LetterDiary | 31 | 10 | 3 | 6 | 0 | 12 | 0 | 28 | 3 | 1 | 0.90 | 0.95 |
| Library | 3 | 3 | 0 | 0 | 0 | 0 | 0 | 3 | 0 | 1 | 1.00 | 1.00 |
| Linguistic | 14 | 1 | 1 | 7 | 1 | 3 | 1 | 11 | 3 | 1 | 0.79 | 0.88 |
| Literature | 11 | 8 | 0 | 0 | 1 | 1 | 1 | 9 | 2 | 1 | 0.82 | 0.90 |
| Logic | 4 | 0 | 0 | 2 | 2 | 0 | 0 | 2 | 2 | 1 | 0.50 | 0.67 |
| Math | 7 | 1 | 1 | 1 | 0 | 3 | 1 | 5 | 2 | 1 | 0.71 | 0.83 |
| Medicine | 35 | 20 | 7 | 6 | 0 | 1 | 1 | 27 | 8 | 1 | 0.77 | 0.87 |
| Novel | 108 | 25 | 6 | 59 | 2 | 13 | 3 | 97 | 11 | 1 | 0.90 | 0.95 |
| Music | 2 | 0 | 0 | 0 | 0 | 0 | 2 | 0 | 2 | 1 | 0.00 | 0.00 |
| Other | 10 | 0 | 1 | 0 | 0 | 9 | 0 | 9 | 1 | 1 | 0.90 | 0.95 |
| Physics | 2 | 1 | 0 | 0 | 0 | 1 | 0 | 2 | 0 | 1 | 1.00 | 1.00 |
| PlayDrama | 12 | 4 | 1 | 2 | 3 | 2 | 0 | 8 | 4 | 1 | 0.67 | 0.80 |
| Pol Sc | 5 | 3 | 0 | 0 | 1 | 1 | 0 | 4 | 1 | 1 | 0.80 | 0.89 |
| Psychology | 6 | 2 | 2 | 0 | 0 | 2 | 0 | 4 | 2 | 1 | 0.67 | 0.80 |
| Religion | 7 | 3 | 0 | 2 | 1 | 1 | 0 | 6 | 1 | 1 | 0.86 | 0.92 |
| Scientific | 15 | 4 | 0 | 3 | 1 | 4 | 3 | 11 | 4 | 1 | 0.73 | 0.85 |
| Sculpture | 1 | 0 | 0 | 0 | 0 | 1 | 0 | 1 | 0 | 1 | 1.00 | 1.00 |
| Sociology | 0 | 0 | 0 | 0 | 0 | 0 | 0 | 0 | 0 | 0 | NA | NA |
| Textbook | 29 | 6 | 2 | 3 | 0 | 16 | 2 | 25 | 4 | 1 | 0.86 | 0.93 |
| TranslatedLit | 18 | 8 | 1 | 3 | 2 | 4 | 0 | 15 | 3 | 1 | 0.83 | 0.91 |
| Vetenary | 4 | 2 | 2 | 0 | 0 | 0 | 0 | 2 | 2 | 1 | 0.50 | 0.67 |
| Zoology | 54 | 24 | 8 | 12 | 0 | 10 | 0 | 46 | 8 | 1 | 0.85 | 0.92 |
| ShortStory | 128 | 50 | 8 | 37 | 3 | 29 | 1 | 116 | 12 | 1 | 0.91 | 0.95 |
| MassMedia | 809 | 247 | 54 | 232 | 51 | 170 | 55 | 649 | 160 | 1 | 0.80 | 0.89 |
| **OVER ALL** | **1747** | **587** | **129** | **483** | **71** | **397** | **80** | **1467** | **280** | **1** | **0.84** | **0.91** |

We have also used a JAVA inbuilt function to handle all types of morphologically derived forms of মাথা (māthā) available in the corpus, like, মাথায় (māthāy), মাথার (māthār), মাথাব্যাথা (māthābyāthā),

মাথাপিছু (māthāpichu), মাথাতে (māthāte), etc. We have achieved 100% accuracy in this regard, which results the Precision of the output is 1. We have achieved over all 84% accuracy in case of 1747 sentences. In most of the cases the output is satisfactory, in the sense that it has rightly referred to the intended sense. However, in certain cases, the performance of the system is not up to the mark as it failed to capture the actual sense of the word used in specific syntactic frame. We have looked into each distinct case of failure and investigated the results closely to identify the pitfalls of the results (Section 8).

## 8. FEW CLOSE OBSERVATIONS

The following parameters mattered the most on the output during the execution of the system:

• As prior probability of each class (P(ci)) depends on the number of sentences in each class, probability is affected due to huge difference between the number of sentences in each class. To overcome this, it is wiser to keep number of sentences in each category constant (55 in our case).

• As the total number of word frequency of each class (ni) remains at delimiter, the conditional probability of keyword occurrence in a given class is affected due to the huge inequality between the total number of word frequency in each class. But this incidence is not under any one's control, because the input sentences are taken from a real life data set. For this reason, results in few cases are derived wrong.

• Use of dissimilar Bengali fonts has a very adverse impact on the output. Text data needs to be rendered in uniform font.

• At certain cases, irrelevant words used in a sentence have caused inconsistency in final calculation. These sentence structures matter a lot on the accuracy of the results. As for example:

a) স্নানার্থিনীর বসে থাকার ভঙ্গীটি স্বাভাবিকতায় অনবদ্য মসৃণ ও আয়ত পিঠ বুক পেছন থেকে স্বভাবতই অদৃশ্য বাঁহাত কাপড়ে ঢাকা ও সামনের দিকে মাঝখান থেকে এগোনো ডান হাতটি একটু বেঁকে আঙুলগুলি নরম বসার জায়গাটি ছুঁয়েছে বসার জায়গায় সাদা কাপড়ের ভাঁজ প্রান্তের নাতি সূক্ষ্ম নীলচে রঙের সূচীকর্ম মাথায় প্রাকস্নানের কাপড় দিয়ে চুলের পেছন দিক ঢাকা ঈষত্ ডানদিকে ঘোরানো মুখাবয়বে শুধু নাক ও চোখের পাতার দৃশ্য এবং স্থির।

b) চারা জন্মানোর পর চারা ওঠার পর ২০ ৩০ দিন পরে হেক্টর প্রতি ৪০০ লিটার জলে ৭.৫ লিটার আনসার ৫২৯ মিশাইয়া পাট গাছের সারির মধ্যে স্প্রেয়ার যন্ত্রের মাথায় ঘোরা টোপ ঢাকনা লাগাইয়া এমন ভাবে স্প্রে করিতে হইবে যেন শুধু আগাছার গায়ে ঔষধ লাগে। etc.

• Another major problem is just the opposite of the issue stated above. Here the short length of a sentence (with regard to number of words) is a hindrance in capturing the intended meaning of the word. As for example: হ্যাঁ আমি মাথা নাড়ি। ও আচ্ছা আচ্ছা আমি মাথা দোলাই। etc. In case of such sentences, after discarding the stop words, there remains insufficient information into the sentences to sense the actual meaning of the ambiguous word.

We have tracked as well as analyzed all the 280 wrong outputs and noted that most of these have occurred due to very long or very short syntactic constructions.

## 9. CONCLUSION AND FUTURE WORKS

In this paper, with a single lexical example, we have tried to show that our proposed approach can disambiguate the sense of an ambiguous word. Except few cases, the result obtained from our system is quite satisfactory according to our expectation. We argue that a stronger and properly populated learning set would invariable yield better result. In future, we plan to disambiguate the most frequently used 100 ambiguous words of different parts-of-speech in Bengali.

Finally, as the target word used in our experiment is a noun, it is comparatively easy to handle its inflections as the number of inflected forms is limited. In case of a verb, however, we may need a stemmer for retrieving the stem or lemma from a conjugated verb form.

## Authors

Alok Ranjan Pal has been working as an a Assistant Professor in Computer Science and Engineering Department of College of Engineering and Management, Kolaghat since 2006. He has completed his Bachelor's and Master's degree under WBUT. Now, he is working on Natural Language Processing.

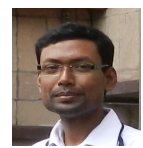

Dr. Diganta Saha is currently working as an Associate Professor in Department of Computer Science & Engineering, Jadavpur University, Kolkata. His field of specialization is Machine Translation/ Natural Language Processing/ Mobile Computing/ Pattern Classification.

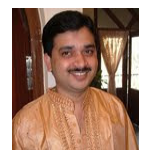

Dr. Niladri Sekhar Dash is currently working as an Assistant Professor in Department of Linguistics and Language Technology, Indian Statistical Institute, Kolkata. His field of specialization is Corpus Linguistics, Natural Language Processing, Language Technology, Word Sense Disambiguation, Computer Assisted Language Teaching, Machine Translation, Computational Lexicography, Field Linguistics, Graded Vocabulary Generation, Applied Linguistics, Educational Technology, Bengali Language and Linguistics.

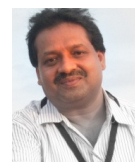